\documentclass{article}

\usepackage{microtype}
\usepackage{graphicx}
\usepackage{subcaption}
\usepackage{booktabs} 

\usepackage{hyperref}

\usepackage[accepted]{icml2026}

\usepackage{hyperref}

\usepackage{url}
\usepackage{xcolor}         
\usepackage[noend]{algpseudocode}
\usepackage{xurl}      

\usepackage{graphicx}
\usepackage{amsmath}
\usepackage{cleveref}
\usepackage{amsfonts}       
\usepackage{colortbl}
\usepackage{multirow}
\usepackage{graphicx}
\usepackage{subcaption}
\usepackage[table]{xcolor}
\usepackage{tabularx}  
\usepackage{amsthm}      
\usepackage{enumitem}    
\usepackage{wrapfig} 
\usepackage{colortbl} 
\usepackage{adjustbox}
\usepackage{array}
\usepackage{amssymb}
\usepackage{tcolorbox}
\tcbuselibrary{listings,skins,breakable,listingsutf8}
\usepackage[scaled=0.92]{inconsolata} 
\usepackage{listings}
\newcolumntype{Y}{>{\centering\arraybackslash}X}

\definecolor{codegray}{rgb}{0.5,0.5,0.5}
\definecolor{codepurple}{rgb}{0.58,0,0.82}
\definecolor{backcolour}{rgb}{0.95,0.95,0.95}

\lstdefinelanguage{json}{
  basicstyle=\ttfamily\small,
  numbers=left,
  numberstyle=\tiny\color{codegray},
  stepnumber=1,
  numbersep=5pt,
  showstringspaces=false,
  breaklines=true,
  frame=single,
  backgroundcolor=\color{backcolour},
  stringstyle=\color{codepurple},
  morestring=[b]",
  literate=
   *{0}{{{\color{black}0}}}{1}
    {1}{{{\color{black}1}}}{1}
    {2}{{{\color{black}2}}}{1}
    {3}{{{\color{black}3}}}{1}
    {4}{{{\color{black}4}}}{1}
    {5}{{{\color{black}5}}}{1}
    {6}{{{\color{black}6}}}{1}
    {7}{{{\color{black}7}}}{1}
    {8}{{{\color{black}8}}}{1}
    {9}{{{\color{black}9}}}{1}
}

\usepackage{tcolorbox}
\tcbuselibrary{listings}

\newtcolorbox{evalprompt}{
  colback=gray!5!white,
  colframe=black!70,
  fonttitle=\bfseries,
  title=Evaluation Prompt,
  enhanced,
  sharp corners,
  fontupper=\ttfamily\small,
  before upper=\parindent0pt,
}

\newtcolorbox{systemprompt}{
  colback=gray!5!white,
  colframe=black!70,
  fonttitle=\bfseries,
  title=System Prompt,
  enhanced,
  sharp corners,
  fontupper=\ttfamily\small,
  before upper=\parindent0pt,
}

\theoremstyle{plain}

\theoremstyle{definition}

\theoremstyle{remark}

\usepackage[textsize=tiny]{todonotes}

\icmltitlerunning{DeFacto: Counterfactual Thinking with Images for Enforcing Evidence-Grounded and Faithful Reasoning}

\begin{document}

\twocolumn[
  \icmltitle{DeFacto: Counterfactual Thinking with Images for Enforcing Evidence-Grounded and Faithful Reasoning}



    \icmlsetsymbol{equal}{*}
    \icmlsetsymbol{corr}{\dagger}
    \begin{icmlauthorlist}
      \icmlauthor{Tianrun Xu}{tsinghua,zgc}
      \icmlauthor{Haoda Jing}{tsinghua}
      \icmlauthor{Ye Li}{xinjiang}
      \icmlauthor{Yuquan Wei}{fuzhou}
      \icmlauthor{Jun Feng}{zgc,cas}
      \icmlauthor{Guanyu Chen}{tsinghua}
      \icmlauthor{Haichuan Gao}{qianjue}
      \icmlauthor{Tianren Zhang}{tsinghua}
      \icmlauthor{Jing Liu}{zgc,cas,ucas-ai}
      \icmlauthor{Feng Chen$^{\dagger}$}{tsinghua}
    \end{icmlauthorlist}
    \icmlaffiliation{tsinghua}{Department of Automation, Tsinghua University, Beijing, China}
    \icmlaffiliation{zgc}{Zhongguancun Academy, Beijing, China}
    \icmlaffiliation{xinjiang}{School of Software, Xinjiang University, Urumqi, China}
    \icmlaffiliation{fuzhou}{College of Materials Science and Engineering, Fuzhou University, Fuzhou, China}
    \icmlaffiliation{cas}{Institute of Automation, Chinese Academy of Sciences, Beijing, China}
    \icmlaffiliation{ucas-ai}{School of Artificial Intelligence, University of Chinese Academy of Sciences, Beijing, China}
    \icmlaffiliation{qianjue}{Beijing Qianjue Technology Co., Ltd., Beijing, China}
    \icmlcorrespondingauthor{Feng Chen}{chenfeng@mail.tsinghua.edu.cn}
    \icmlkeywords{Machine Learning, ICML}

  \vskip 0.3in
]



\printAffiliationsAndNotice{}  

\begin{abstract}
Recent advances in multimodal language models (MLLMs) have made \emph{thinking with images} a dominant paradigm for multimodal reasoning. However, existing methods still fail to ensure \emph{evidence--answer consistency}, where correct answers must be supported by correct visual evidence. To address this issue, we propose \textit{DeFacto}, a counterfactual reasoning framework that explicitly aligns visual evidence with final answers. Our approach integrates three complementary training paradigms: (i) positive, (ii) counterfactual, and (iii) random-masking. We further develop a language-guided evidence construction pipeline that automatically localizes question-relevant regions and generates counterfactual variants, resulting in \textbf{DeFacto-100K}. Building on this dataset, we train MLLMs with GRPO-based reinforcement learning and design three complementary rewards to promote correct answering, structured reasoning, and consistent evidence selection. Moreover, we introduce \textbf{DeFacto-1.5K}, a human-annotated benchmark for systematically evaluating evidence-grounded consistency beyond answer accuracy. Experiments on diverse benchmarks demonstrate that \textit{DeFacto} substantially improves both answer accuracy and evidence--answer consistency over strong baselines. The code and datasets are available at \url{https://github.com/tinnel123666888/defacto}.
\end{abstract}

\section{Introduction}

Vision-language models (VLMs)~\cite{alayrac2022flamingo,li2023blip,zhu2023minigpt,liu2023visual,liu2024improved,peng2023kosmos,bai2023qwen,team2023gemini,chen2024far,xu2025ouro,yan2026act} have achieved remarkable progress in recent years, demonstrating strong capabilities across a wide range of multimodal tasks such as visual question answering, image captioning, and referring expression comprehension. By leveraging large-scale pretraining and cross-modal alignment, these models can generate fluent and semantically relevant outputs grounded in visual context. However, in complex scenarios that require multi-step reasoning or fine-grained perception, existing models often rely heavily on implicit language priors, producing plausible yet unfaithful responses that are weakly grounded in the actual image. Instead of genuinely learning to reason over visual content, these models often fall back on text-based chain-of-thought patterns, limiting their ability to handle cases where critical evidence must be directly perceived from the image.

\begin{figure*}[t]
    \centering
    \includegraphics[width=0.99\linewidth]{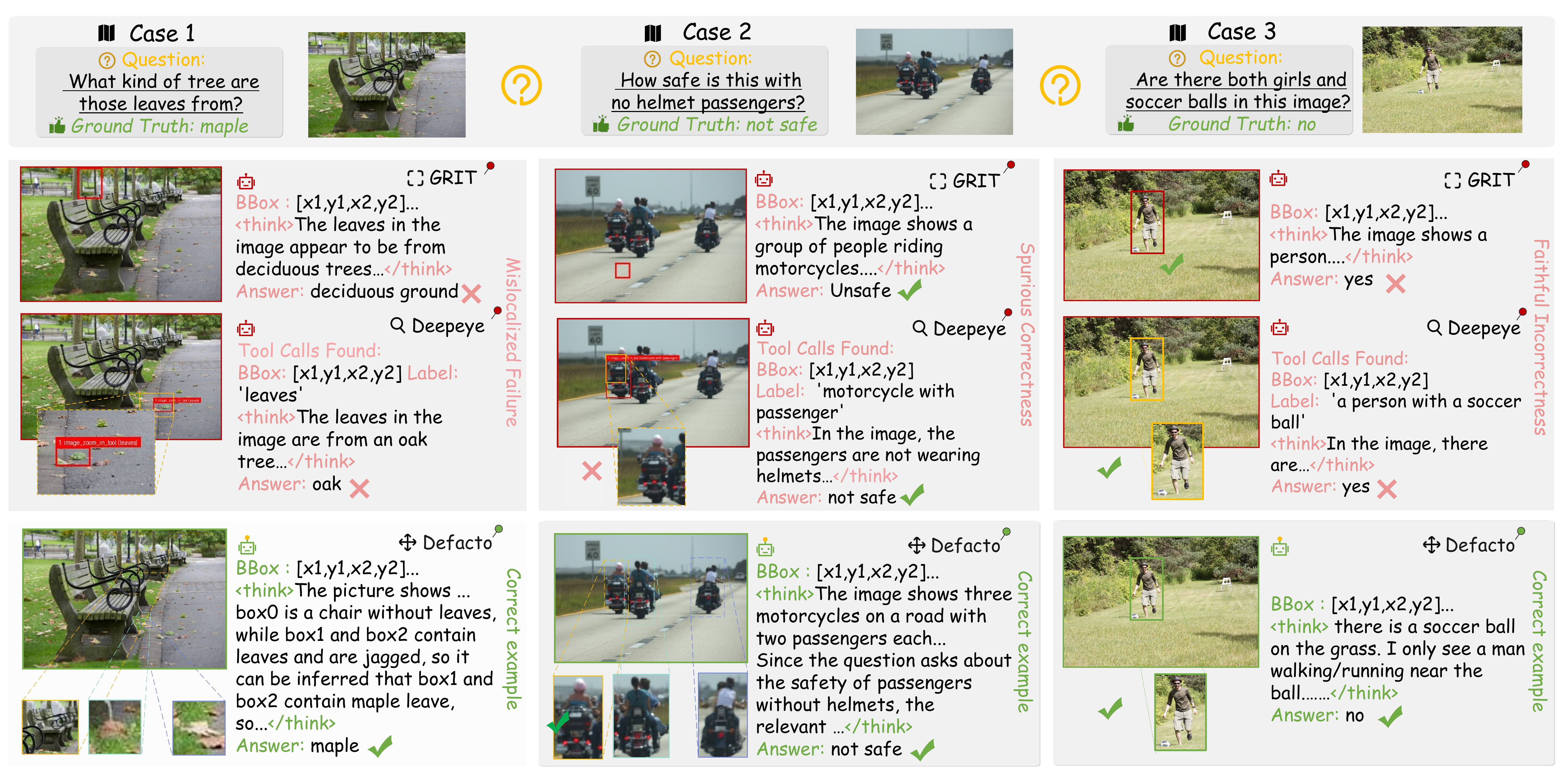}
    \caption{Qualitative examples of \textbf{unfaithful reasoning} in thinking-with-images. We categorize failures into three types: \emph{Mislocalized Failure} (wrong evidence, wrong answer), \emph{Spurious Correctness} (wrong evidence, correct answer), and \emph{Faithful Incorrectness} (correct evidence, wrong answer). We compare two representative baselines (Deepeyes and GRIT) with our method on three cases.}
    \label{fig:qual_cases}
\end{figure*}

Recent advances in “thinking with images”~\cite{gpt4o_20240806,openai2025thinking} integrate explicit visual steps into multimodal reasoning to improve interpretability and visual grounding. Early approaches adopt supervised fine-tuning (SFT)~\cite{ouyang2022training,touvron2023llama,liu2023visual,dettmers2023qlora,xu2025multi}, where models learn to generate region-aware reasoning traces in a chain-of-thought manner with manually annotated visual steps~\cite{shao2024visual}. To reduce annotation cost, subsequent works explore reinforcement learning to enable autonomous visual interactions such as cropping, attention shifting, and zooming~\cite{zheng2025deepeyes,cao2025ground,liu2025visual,zhang2025chain,pixelreasoner}. Yet existing thinking-with-images methods still cannot guarantee \emph{evidence--answer consistency}, often exhibiting three failure cases: (i) \textbf{Mislocalized Failure} (wrong evidence, wrong answer), (ii) \textbf{Spurious Correctness} (wrong evidence, correct answer), and (iii) \textbf{Faithful Incorrectness} (correct evidence, wrong answer). As shown in Fig.~\ref{fig:qual_cases}, we use two representative baselines, Deepeyes and GRIT, to qualitatively illustrate these issues. These observations suggest that reliable multimodal reasoning requires jointly enforcing both correct answering and evidence-aligned visual reasoning, rather than optimizing accuracy alone.

Motivated by these failure cases, we introduce \textit{DeFacto}, a counterfactual reasoning framework that aligns reasoning trajectories with visual evidence to improve evidence--answer consistency. The core idea is to jointly constrain the model via three complementary training forms: (i) \textbf{positive supervision}, (ii) \textbf{counterfactual abstention}, and (iii) \textbf{random masking}. In the positive case, the model is trained on the original image to predict evidence regions $\mathcal{R}^+$ (bounding boxes covering essential visual cues) and the correct answer, receiving positive feedback only when both are correct. In the counterfactual case, the regions $\mathcal{R}^+$ are masked and the model is required to abstain by outputting a designated token (e.g., “unknown”). In the random-masking case, irrelevant regions $\mathcal{R}^-$ are masked to prevent shortcut learning and ensure the abstention behavior is triggered only by missing evidence rather than the presence of masks.

In practice, constructing counterfactual samples requires reliably identifying the question-relevant regions. To this end, we adopt a structured two-stage extraction pipeline. First, a multimodal language model (Qwen2.5-VL~\cite{Qwen2.5-VL}) parses the question and generates a set of key descriptors (e.g., “the red cup,” “the text on the shirt”). Next, candidate regions in the image are obtained from a region proposal network (RPN)~\cite{ren2015faster} and an OCR module~\cite{islam2017survey}. The OCR regions are further matched with textual descriptors to capture evidence critical for text-centric questions. For visual objects, the descriptors are fed into an open-vocabulary detector (DINO-X~\cite{ren2024dinoxunifiedvisionmodel}), which provides bounding boxes that serve as positive evidence regions. Finally, the remaining proposals from the RPN, after removing matched positives, are treated as irrelevant regions for counterfactual construction. Using this pipeline, we construct \textbf{DeFacto-100K}, a counterfactual dataset of about 100k images, ensuring that positive, counterfactual, and random-masking instances differ only in the availability of essential evidence while preserving unrelated context. Building on \textbf{DeFacto-100K}, the model is further optimized with GRPO-based reinforcement learning. This training paradigm enforces consistency between evidence selection and final predictions, ensuring that reasoning traces remain aligned with the visual evidence used to justify the answer. As illustrated in Fig.~\ref{fig:qual_cases}, our method selects regions that support the question and maintains coherent evidence-to-answer consistency.

Our main contributions are summarized as follows:

(1) We propose \textit{DeFacto}, a counterfactual \emph{thinking with images} framework that aligns the reasoning process with essential visual evidence by jointly optimizing answer correctness and region-level faithfulness via GRPO-based reinforcement learning.

(2) We construct \textbf{DeFacto-100K}, a counterfactual dataset of about 100k images built with a language-guided pipeline that integrates open-vocabulary detection with targeted evidence masking, ensuring that only question-relevant regions are removed while irrelevant context is preserved.

(3) We introduce \textbf{DeFacto-1.5K}, a 1.5k-sample faithfulness benchmark with human-verified evidence annotations, enabling systematic evaluation of region-level reasoning faithfulness beyond answer accuracy.

(4) Through extensive experiments on diverse VQA benchmarks as well as \textbf{DeFacto-1.5K}, we demonstrate that \textit{DeFacto} consistently improves both answer accuracy and grounding faithfulness over strong baselines.

\paragraph{Conflict of Interest Disclosure.}
The authors declare no financial conflicts of interest related to this work.

\section{Related Work}

\textbf{Multimodal Large Language Models.}
Recent years have witnessed rapid progress in multimodal large language models (MLLMs), which extend large language models with visual encoders to support unified vision--language understanding and reasoning. Representative proprietary systems such as GPT-4V and GPT-4o~\cite{gpt4o_20240806} demonstrate strong multimodal capabilities and accelerate the trend of integrating explicit visual interaction into reasoning~\cite{openai2025thinking}. On the open-source side, Qwen-VL/Qwen2.5-VL~\cite{Qwen2.5-VL} provides scalable and strong multimodal backbones with competitive OCR and document understanding; LLaVA-style models~\cite{liu2023visual} popularize instruction-tuned MLLMs by aligning vision encoders with LLMs through multimodal instruction data;  and DeepSeek-VL~\cite{wu2024deepseek} advances large-scale multimodal pretraining with competitive performance.

\textbf{Counterfactual Reasoning in Vision-Language Models.}  
Counterfactual reasoning in VLMs can be categorized into two types: counterfactual data generation and inference-based reasoning. The first enhances robustness by constructing or augmenting counterfactual samples to reduce bias and hallucination. For example, Learning Chain of Counterfactual Thought~\cite{zhang2020counterfactual} disentangles factual knowledge from reasoning via CoBRa and CoCT datasets; C-VQA~\cite{zhang2024if} and CRIPP-VQA~\cite{patel2022cripp} construct benchmarks for counterfactual VQA in static and video settings, respectively; Counterfactual Vision and Language Learning~\cite{abbasnejad2020counterfactual} and Counterfactual Contrastive Learning~\cite{zhang2024learning} generate counterfactuals through structural causal models and perturbation strategies, while CounterCurate~\cite{zhang2024countercurate} improves compositional reasoning by augmenting training data with physically grounded examples and semantic counterfactuals using generative models. The second type focuses on inference with mechanisms such as Counterfactual-based Saliency Maps~\cite{wang2023counterfactual} for contrastive visual explanation, DiG-IN~\cite{augustin2024dig} for diffusion-guided latent edits, and Counterfactual VQA~\cite{niu2021counterfactual} for causal effect modeling. CFMM~\cite{li2024eyes} curates a manually labeled benchmark of presupposition-based counterfactual questions to systematically expose MLLMs' limitations in hypothetical visual reasoning. Beyond counterfactual perturbations, MM-Verify~\cite{sun2025mmverifyenhancingmultimodalreasoning} addresses a related failure mode in multimodal reasoning by introducing a verification model that checks chain-of-thought trajectories.

\textbf{Structured Thinking with Images in Vision-Language Models.}  
The concept of "thinking with images" was initially highlighted in OpenAI o3~\cite{achiam2023gpt,openai2025thinking} and later explored in works like COGCOM~\cite{qi2024cogcom} and GRIT~\cite{fan2025grit}. Recent datasets also highlight the importance of evaluating visual reasoning beyond raw perception. For example, VisCoT~\cite{shao2024visual} provides visual-evidence for vqa, while datasets such as MSTI~\cite{chen2024cofipara} emphasize structured visual understanding across detection, entity grounding, and answer generation. Existing approaches can be broadly categorized into two classes. The first category includes GRIT, which combines natural language and bounding boxes via reinforcement learning; REFOCUS~\cite{fu2025refocus}, which formulates visual editing as intermediate reasoning steps; COGCOM~\cite{qi2024cogcom}, which models reasoning as visual manipulations such as cropping and OCR; and VisionReasoner~\cite{liu2025visionreasoner}, which unifies detection, segmentation, and counting under one framework. The second category emphasizes grounding quality. Fast-and-Slow Visual Agents~\cite{sun2024visual} model dual-system reasoning. DeepEyes~\cite{zheng2025deepeyes} leverages reinforcement learning to train multimodal chains-of-thought and dynamically invoke zoom-in tools when visual evidence is ambiguous. MLLMs Know Where to Look~\cite{zhang2025mllms} improves small-object perception by applying inference-time cropping strategies to highlight fine details. Chain-of-Focus~\cite{zhang2025chain} further adapts zoom-in operations through reinforcement learning, enabling multi-scale reasoning across cluttered scenes. Ground-R1~\cite{cao2025ground} enhances faithfulness by introducing explicit reward signals that align reasoning outputs with grounded evidence. V*~\cite{wu2024v} formulates guided visual search as a core cognitive mechanism to explore high-resolution images efficiently. Visual-RFT~\cite{liu2025visual} refines grounding via reinforcement fine-tuning. PAPO~\cite{wang2025perception} introduces an implicit perception loss into GRPO to jointly optimize perception and reasoning. TreeVGR~\cite{wang2025traceable} explicitly supervises bounding-box generation with a dual IoU reward during RL, providing traceable evidence for visual grounded reasoning.

\begin{figure*}[t]
  \centering
  \includegraphics[width=\textwidth]{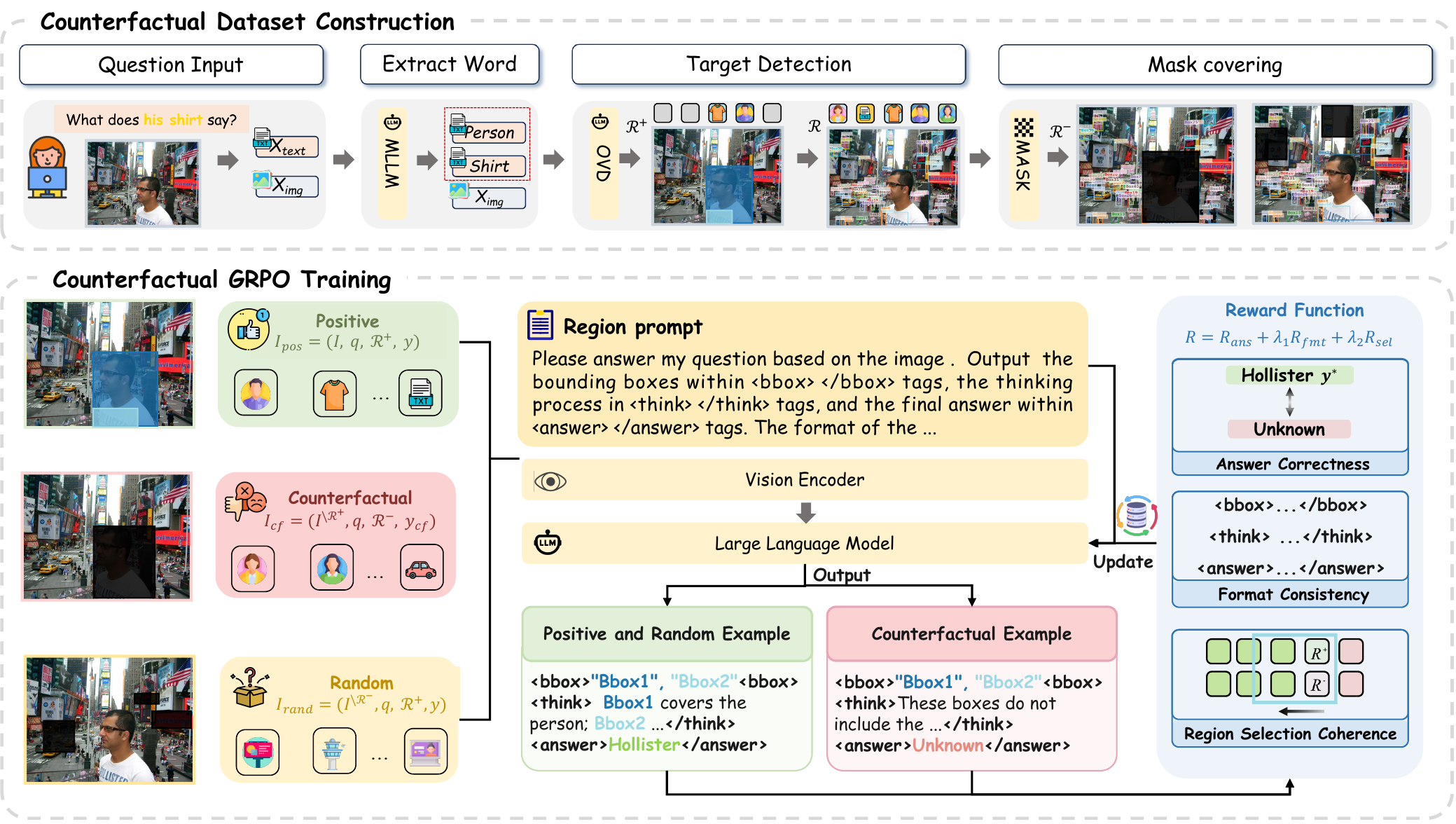}
  \caption{An overview of our counterfactual framework with three inputs: positive (full evidence), counterfactual (masked evidence), and random (masked irrelevant regions).}
  \label{fig:pipeline}
\end{figure*}

\section{Method}
In this section, we present the overall framework of \textsc{DeFacto}, our counterfactual ``thinking with images'' approach. This section is organized into three parts: (1) the overall architecture and inference pipeline (Section~\ref{sec:framework}); (2) the construction of counterfactual datasets via region masking and open-vocabulary filtering (Section~\ref{sec:counterfactual_dataset}); and (3) the reinforcement learning strategy with a tailored reward design that guides the model toward accurate answering, faithful reasoning, and their consistency (Section~\ref{sec:training}).

\subsection{Overall Framework}
\label{sec:framework}

\textsc{DeFacto} is a vision-language reasoning framework that enforces region-level faithfulness in multimodal question answering. It is designed to teach models not only \emph{where to look} in the image but also \emph{when to abstain} if the necessary evidence is absent. By combining structured prompting with counterfactual supervision, \textsc{DeFacto} aligns the reasoning process with visual evidence rather than spurious correlations.

As illustrated in Figure~\ref{fig:pipeline}, given an image and a question explicitly specified in the prompt, the model is required to generate outputs in a unified structured format composed of three fields.
The \texttt{<bbox>} field encodes one or multiple evidence regions as a JSON list of \texttt{[x1, y1, x2, y2]} coordinates, or outputs \texttt{null} when no valid visual grounding is available.
The \texttt{<think>} field provides a concise reasoning process that must be explicitly grounded in the selected bounding boxes (if any) and directly related to the given question.
The \texttt{<answer>} field delivers the final prediction. This structured format ensures that every reasoning trajectory is explicitly tied to visual evidence through bounding boxes and aligned with the model’s final answer. Training is based on three complementary supervision forms. In the positive case, evidence-bearing regions remain visible and the model is rewarded for selecting them and producing the correct answer. In the counterfactual case, these regions are masked, and the model is expected to abstain by outputting \texttt{unknown}. In the random-masking case, irrelevant regions are occluded to prevent shortcut learning from superficial mask patterns. Together, these three forms establish a consistent learning signal that requires both the reasoning path and the answer to be faithful to the underlying visual support.

\subsection{Counterfactual Dataset Construction}
\label{sec:counterfactual_dataset}

\paragraph{Positive, Counterfactual, and Random Instances.}  
Let $\mathcal{R}=\{r_1, r_2, \dots, r_n\}$ denote the set of candidate regions in an image $I$, obtained from a region proposal network (RPN)~\cite{ren2015faster} together with OCR to cover both object-level and text-bearing regions. Among them, $\mathcal{R}^+ \subseteq \mathcal{R}$ represents the evidence regions that are critical to answering the question $q$, while $\mathcal{R}^- = \mathcal{R}\setminus \mathcal{R}^+$ denotes the remaining irrelevant regions. Based on these definitions, we construct three complementary training instances:

\begin{equation}
\begin{aligned}
I_{\text{pos}}  &= (I, q, \mathcal{R}^+, y), \\
I_{\text{cf}}   &= (I^{\setminus \mathcal{R}^+}, q, \mathcal{R}^-, y_{\text{cf}}), \\
I_{\text{rand}} &= (I^{\setminus \mathcal{R}^-}, q, \mathcal{R}^+, y).
\end{aligned}
\label{eq:instances}
\end{equation}

where $y$ is the ground-truth answer, $I_{\text{pos}}$ is the positive instance with evidence regions available, $I_{\text{cf}}$ is the counterfactual instance where evidence regions are masked and the abstention label $y_{\text{cf}}$ (e.g., ``Unknown'') is required, and $I_{\text{rand}}$ is the random-masking instance where irrelevant regions are occluded to prevent shortcut learning.

\paragraph{Construction Process.}  
To automatically construct $I_{\text{pos}}$, $I_{\text{cf}}$, and $I_{\text{rand}}$ without manual annotations, we follow three steps:  

(1) Descriptor extraction.  
Given an image $I$ and a question $q$, we employ a MLLM (Qwen2.5-VL~\cite{Qwen2.5-VL}) to extract a set of key descriptors:
\begin{equation}
\operatorname{MLLM}(I,q) = \mathcal{K}(q) = \{d_1, d_2, \dots, d_m\},
\label{eq:descriptor}
\end{equation}
where each $d_i$ is a textual phrase (e.g., an object, attribute, or relation) that captures the visual concepts in $I$ explicitly mentioned or implied by $q$.  
As illustrated in Fig.~\ref{fig:pipeline} (``Automate Evidence Masking''), for the question ``What does his shirt say?’’, the MLLM decomposes the query into descriptors such as ``a man’’ and ``man’s shirt’’ as the critical evidence.

(2) Evidence localization.  
Let $\mathcal{R}=\{r_1,\dots,r_n\}$ be the set of candidate image regions.  
We employ the open-vocabulary detector DINO-X~\cite{ren2024dinoxunifiedvisionmodel}, which computes grounding scores $\mathsf{OVD}(r,k)$ for each $r\in\mathcal{R}$ and $k\in\mathcal{K}(q)$.  
Based on these scores, the regions are partitioned into evidence and irrelevant sets:
\begin{equation}
\begin{aligned}
\mathcal{R}^+ &= \left\{ r \in \mathcal{R} \ \middle|\ 
\max_{k \in \mathcal{K}(q)} \mathsf{OVD}(r, k) > \tau \right\}, \\
\mathcal{R}^- &= \mathcal{R} \setminus \mathcal{R}^+ .
\end{aligned}
\end{equation}

where $\tau$ is a confidence threshold. In the street example, $\mathcal{R}^+$ corresponds to bounding boxes covering the signboard, while $\mathcal{R}^-$ contains all other regions.

(3) Instance generation.  
Once $\mathcal{R}^+$ and $\mathcal{R}^-$ are obtained, the positive, counterfactual, and random-masking instances are directly constructed as defined in Eq.~\ref{eq:instances}.

For counterfactual dataset construction, we leverage a broad collection of visual question answering and document understanding benchmarks, including VQAv2~\cite{goyal2017making} for general visual question answering, ChartQA~\cite{masry2022chartqa} for chart-based reasoning, and DocVQA~\cite{mathew2021docvqa} for document understanding, among others. This diverse coverage ensures that counterfactual supervision is tested across natural images, scientific diagrams, documents, charts, tables, and multi-source reasoning tasks.

\subsection{Counterfactual GRPO Training}
\label{sec:training}

\paragraph{Sequential Reasoning Formulation.}  
We formulate the reasoning process of \textsc{DeFacto} as a Markov Decision Process (MDP), where the model interacts with the question and image in a sequential manner. At each step, the state $s_t$ encodes the multimodal context, including the input question, the image representation, and the history of previously predicted regions. The policy $\pi_\theta$ then outputs either a new bounding box that localizes question-relevant evidence or a special \texttt{STOP} token to terminate the process.  

Formally, the state at step $t$ is defined as
\begin{equation}
s_t = \{ q, f_v(I), B_{<t} \},
\end{equation}
where $q$ is the question, $f_v(I)$ the image representation, and $B_{<t}$ the set of bounding boxes predicted before step $t$. The rollout continues until \texttt{STOP} is emitted or the maximum step limit is reached, and the final answer is generated based on the accumulated trajectory.

\paragraph{Reward Design.}  
To make training effective, we design three reward components.  
(1) \emph{Answer Correctness Reward}: encourages correct answers in positive/random cases, rewards ``Unknown'' in counterfactual cases, and penalizes unsupported guesses.  
(2) \emph{Format Consistency Reward}: ensures outputs strictly follow the required schema.  
(3) \emph{Region Selection Coherence Reward}: promotes overlap with evidence regions $\mathcal{R}^+$ and penalizes overlap with irrelevant regions $\mathcal{R}^-$, with no reward in counterfactual cases.

The overall training signal combines these components into the composite reward in Eq.~\ref{eq:allr}.
\begin{equation}
R \;=\; R_{\text{ans}} \;+\; \lambda_1\,R_{\text{fmt}} \;+\; \lambda_2\,R_{\text{sel}},
\label{eq:allr}
\end{equation}

\textbf{1. Answer Correctness Reward.}
To enforce correct behavior across the three training forms, we define
\begin{equation}
\label{eq:rans}
\resizebox{\linewidth}{!}{$\displaystyle
R_{\text{ans}}=
\begin{cases}
\mathsf{acc}(\hat y,y^\ast)
- \underbrace{\gamma_{\text{unk}}\,\mathsf{unk}(\hat y)}_{\text{\scriptsize penalize ``Unknown''}},
& t \in \{\text{pos},\text{rand}\}, \\[10pt]
\underbrace{\rho_{\text{unk}}\,\mathsf{unk}(\hat y)}_{\text{\scriptsize reward ``Unknown''}}
- \underbrace{\gamma_{\text{guess}}\,[1-\mathsf{unk}(\hat y)]}_{\text{\scriptsize penalize guess}}
- \underbrace{\gamma_{\text{corr}}\,\mathbf{1}[\hat y=y^\ast]}_{\text{\scriptsize penalize even if correct}},
& t=\text{cf}.
\end{cases}
$}
\end{equation}

where $\mathsf{acc}(\hat y,y^\ast)\!\in\!\{0,1\}$ indicates answer correctness, and $\mathsf{unk}(\hat y)\!\in\!\{0,1\}$ indicates an ``Unknown'' response. Here $\gamma_{\text{unk}}>0$ penalizes answering ``Unknown'' in positive or random cases, $\rho_{\text{unk}}>0$ rewards ``Unknown'' in counterfactual cases, $\gamma_{\text{guess}}>0$ penalizes any concrete guess in counterfactual cases, and $\gamma_{\text{corr}}>\gamma_{\text{guess}}$ applies an even stronger penalty when the model outputs the correct answer $y^\ast$ without access to evidence.

{\bfseries 2. Format Consistency Reward.} We encourage well-formed outputs with valid bounding box coordinates and the required response structure:
\begin{equation}
R_{\text{fmt}} \;=\;
\begin{cases}
\;\;\alpha, & \text{if output follows the required schema},\\
\;\;0, & \text{otherwise}.
\end{cases}
\end{equation}
Here, the ``required schema'' refers to the presence of \texttt{<think>}...\texttt{</think>} for the reasoning process, \texttt{<bbox>}...\texttt{</bbox>} for the predicted bounding boxes, and \texttt{<answer>}...\texttt{</answer>} for the final answer. In particular, the \texttt{<bbox>} field must contain well-formed bounding box coordinates in the format \texttt{[x1, y1, x2, y2]}, ensuring that the model explicitly grounds its predictions on localized visual regions.

\begin{table*}[h]
\centering
\caption{Results on General VQA Benchmarks (accuracy, \%). $\Delta$ indicates improvements of DeFacto over Qwen2.5-VL 7B.}
\label{tab:gen_vqa}
\footnotesize
\setlength{\tabcolsep}{5pt}
\renewcommand{\arraystretch}{1.05}
\begin{tabularx}{\textwidth}{@{}l l *{6}{>{\centering\arraybackslash}X}@{}}
\toprule
\textbf{Model} & \textbf{Backbone} & VQAv2 & OKVQA & GQA & SciQA & VizWiz & VSR \\
\midrule
Qwen2.5-VL      & Qwen2.5-VL-7B & \underline{67.0} & \underline{58.9} & \underline{53.2} & 83.0 & 54.1 & 63.6 \\
ViCrop          & LLaVA-1.5 (Vicuna-7B) & --   & --   & --   & 82.1 & \textbf{64.7} & --   \\
GRIT            & Qwen2.5-VL-3B & 64.7 & 43.1 & \underline{54.8} & 59.8 & 39.4 & 62.4 \\
DeepEyes        & Qwen2.5-VL-7B & --   & 40.0 & 45.5 & 49.7 & 32.2 & 43.5 \\
Visual-SR1      & Qwen2.5-VL-7B & 64.7 & 49.0 & 52.3 & \textbf{88.0} & 35.8 & \underline{66.4} \\
\midrule
\rowcolor{gray!20}
DeFacto (Ours)  & Qwen2.5-VL-7B & \textbf{72.1} & \textbf{61.7} & \textbf{63.9} & \underline{83.6} & \underline{61.4} & \textbf{71.0} \\
\rowcolor{gray!20}
$\Delta$ (vs Qwen2.5-VL 7B) & -- & +5.1 & +2.8 & +10.7 & +0.6 & +7.3 & +7.4 \\
\bottomrule
\end{tabularx}
\end{table*}

\begin{table*}[h]
\centering
\caption{Results on Document VQA and Scene Text-centric Benchmarks (accuracy, \%). $\Delta$ indicates improvements of DeFacto over Qwen2.5-VL 7B.}
\label{tab:doc_scene_vqa}
\footnotesize
\setlength{\tabcolsep}{4pt}
\renewcommand{\arraystretch}{1.05}
\begin{tabularx}{\textwidth}{@{}l *{9}{>{\centering\arraybackslash}X}@{}}
\toprule
\textbf{Model}
& \multicolumn{6}{c}{\textbf{Document VQA}}
& \multicolumn{3}{c}{\textbf{Scene Text-centric}} \\
\cmidrule(lr){2-7} \cmidrule(lr){8-10}
& DocVQA & ChartQA & InfoVQA & DeepForm & KLC & WTQ
& STVQA & TextVQA & AI2D \\
\midrule
Qwen2.5-VL
& \underline{92.0} & \underline{74.4} & 71.5 & 27.9 & 36.5 & 62.3
& 67.9 & \underline{79.1} & 69.5 \\
GRIT
& 57.0 & 24.6 & 43.7 & 17.9 & 17.8 & 32.0
& 64.7 & 60.8 & \underline{75.5} \\
ViCrop
& 31.8 & 50.8 & 52.4 & 20.3 & 29.1 & 51.5
& \underline{72.4} & 60.8 & 68.8 \\
DeepEyes
& 67.1 & --   & 39.9 & --   & 32.2 & 51.3
& 45.9 & 40.4 & 37.0 \\
Visual-SR1
& 84.9 & 71.7 & \underline{76.5} & \textbf{50.8} & \underline{36.6} & \textbf{71.3}
& \textbf{75.9} & 70.1 & 69.0 \\
\midrule
\rowcolor{gray!20}
DeFacto (Ours)
& \textbf{94.0} & \textbf{82.1} & \textbf{79.1} & \underline{45.6} & \textbf{38.9} & \underline{63.7}
& 71.2 & \textbf{82.9} & \textbf{76.1} \\
\rowcolor{gray!20}
$\Delta$ (vs Qwen2.5-VL 7B)
& +2.0 & +7.7 & +7.6 & +17.7 & +2.4 & +1.4
& +3.3 & +3.8 & +6.6 \\
\bottomrule
\end{tabularx}
\end{table*}

\textbf{3. Region Selection Coherence Reward.}  
Let $B = \{\text{bbox}^1,\dots,\text{bbox}^k\}$ be the set of bounding boxes predicted before \texttt{STOP}.  
We define the overlap scores with evidence regions $\mathcal{R}^+$ and irrelevant regions $\mathcal{R}^-$ as
\[
\phi^+(b) = \max_{r\in\mathcal{R}^+}\text{IoU}(b,r), 
\quad
\phi^-(b) = \max_{r\in\mathcal{R}^-}\text{IoU}(b,r).
\]

The reward is then defined as
\begin{equation}
\label{eq:rsel}
\resizebox{\linewidth}{!}{$\displaystyle
R_{\text{sel}}=
\begin{cases}
\beta_{\text{pos}}\dfrac{1}{|B|}\sum_{b\in B}\phi^+(b)
- \beta_{\text{neg}}\dfrac{1}{|B|}\sum_{b\in B}\phi^-(b),
& t \in \{\text{pos},\text{rand}\},\ B\neq\varnothing, \\[10pt]
-\gamma_{\varnothing},
& t \in \{\text{pos},\text{rand}\},\ B=\varnothing, \\[6pt]
0,
& t=\text{cf}.
\end{cases}
$}
\end{equation}

with $\beta_{\text{pos}}, \beta_{\text{neg}}, \gamma_{\varnothing} > 0$.

\paragraph{Training Strategy.}  
We fine-tune Qwen2.5-VL with reinforcement learning, using Group Relative Policy Optimization (GRPO) and the composite reward in Eq.~\ref{eq:rans} and Eq.~\ref{eq:rsel}. GRPO compares multiple rollouts within a group and rewards each according to its improvement over the group average, eliminating the need for a value network and reducing variance. The objective is defined as:  
\begin{equation}
\label{eq:grpo}
\resizebox{\linewidth}{!}{$\displaystyle
\mathcal{J}_{\text{GRPO}}(\theta) =
\mathbb{E}_{q,\,\{o_i\}_{i=1}^{G}}\!\left[
\frac{1}{G}\sum_{i=1}^{G}\frac{1}{|o_i|}\sum_{t=1}^{|o_i|}
\Big(
\min\!\big(r_{i,t}(\theta)\,\hat{A}_{i,t},\;
\mathrm{clip}(r_{i,t}(\theta),\,1-\epsilon,\,1+\epsilon)\,\hat{A}_{i,t}\big)
- \beta\,\mathbb{D}_{\mathrm{KL}}\!\left[\pi_\theta\,\|\,\pi_{\text{ref}}\right]
\Big)
\right],
$}
\end{equation}
where the importance ratio is
\begin{equation}
r_{i,t}(\theta) = \frac{\pi_\theta(o_{i,t}\mid q,o_{i,<t})}{\pi_{\theta_{\text{old}}}(o_{i,t}\mid q,o_{i,<t})},
\end{equation}
and the group-relative advantage is computed by normalizing rewards within each sampled group of $G$ responses:
\begin{equation}
\hat{A}_{i,t} = \frac{R(o_i) - \mathrm{mean}(\{R(o_1),\dots,R(o_G)\})}{\mathrm{std}(\{R(o_1),\dots,R(o_G)\})}.
\end{equation}
where we set the group size $G=4$ to balance stability and exploration during training.

\section{Experiment}
\label{sec:experiment}
\subsection{Setup}
\paragraph{Training Configuration.} 
We train all models with the AdamW optimizer using a learning rate of $1\times 10^{-6}$, $(\beta_1,\beta_2)=(0.9,0.999)$, and $\epsilon=1\times 10^{-8}$. Training is performed with a global batch size of 8 and micro-batch size of 1 per GPU, combined with gradient accumulation steps of 2. Gradients are clipped to a maximum norm of 1.0 to ensure stability. We enable BF16 precision training. All experiments are conducted on 8 NVIDIA H100 GPUs with 80GB memory each, and models are trained for one epoch over the collected dataset.

\paragraph{Baselines.} 
We compare \textsc{DeFacto} against a broad set of recent approaches that explicitly incorporate visual reasoning into multimodal language models. Specifically, we include the \textsc{Qwen2.5-VL}~\cite{Qwen2.5-VL}, a strong pretrained backbone widely used for visual understanding; \textsc{ViCrop}~\cite{zhang2025mllms}, which improves small-object perception via inference-time cropping; \textsc{GRIT}~\cite{fan2025grit}, which integrates grounded reasoning traces through reinforcement learning; \textsc{Deepeyes}~\cite{zheng2025deepeyes}, which incentivizes models to call visual tools during reasoning; and \textsc{Visual-SR1}~\cite{li2025self}, which enhances step-by-step visual reasoning with self-refinement. This selection covers both state-of-the-art backbones and recent ``thinking with images'' algorithms for visual reasoning.

\paragraph{Benchmarks.}
Our evaluation spans a diverse collection of visual reasoning benchmarks. For general-purpose VQA, we use OKVQA~\cite{marino2019ok}, VQAv2~\cite{goyal2017making}, GQA~\cite{hudson2019gqa}, VizWiz~\cite{gurari2018vizwiz}, ScienceQA~\cite{lu2022learn}, and VSR~\cite{liu2023visual}. For document- and structure-centric evaluation, we adopt DocVQA~\cite{mathew2021docvqa}, ChartQA~\cite{masry2022chartqa}, InfoVQA~\cite{mathew2022infographicvqa}, DeepForm~\cite{svetlichnaya2020deepform}, Kleister~KLC~\cite{stanislawek2021kleister}, and WikiTableQuestions (WTQ)~\cite{pasupat2015compositional}. To test text-intensive reasoning, we include TextVQA~\cite{singh2019towards}, AI2D~\cite{kembhavi2016diagram}, and STVQA~\cite{biten2019scene}. 
We further construct \textbf{DeFacto-1.5K} by sampling 100 examples from each of the above evaluation benchmarks and manually annotating question-relevant evidence regions. For each question, annotators label one or multiple key bounding boxes that support the correct answer. In addition, we evaluate on more benchmarks, including OCRBench~\cite{liu2024ocrbench}, MMStar~\cite{chen2024we}, MMMU~\cite{yue2024mmmu}, MMB\textsubscript{1.1}~\cite{liu2024mmbench}, and POPE~\cite{li2023evaluating}, where \textsc{DeFacto} achieves competitive performance against both leading closed-source models (e.g., GPT-4o~\cite{gpt4o_20240806}, Gemini-1.5-Pro~\cite{team2024gemini}) and recent thinking-with-images approaches, as summarized in Table~\ref{tab:sota_preview}. 

\begin{table}[htp]
\centering
\caption{Comparison with representative closed-source and thinking-with-images models.}
\label{tab:sota_preview}
\scriptsize
\setlength{\tabcolsep}{4pt}
\renewcommand{\arraystretch}{1.05}
\begin{tabular}{lccccc}
\toprule
\textbf{Model} & \textbf{OCRB} & \textbf{MMStar} & \textbf{MMMU} & \textbf{MMB$_{1.1}$} & \textbf{POPE} \\
\midrule
GPT-4o          & 736 & \textbf{63.9} & \textbf{69.2} & \textbf{82.2} & --   \\
Gemini-1.5-Pro  & \underline{754} & --   & \underline{62.2} & --   & --   \\
\midrule
DeepEyes        & 636 & 43.6 & 44.1 & 29.4 & 87.7 \\
Chain-of-Focus  & 632 & 58.1 & 46.1 & 75.3 & \underline{88.4} \\
Pixel Reasoner  & 597 & 62.9 & 52.5 & 78.5 & 87.8 \\
Visual-SR1      & 449 & 62.8 & 57.2 & 77.4 & 86.0 \\
\midrule
\rowcolor{gray!15}
\textbf{DeFacto (Ours)} & \textbf{871} & \underline{63.2} & 56.6 & \underline{81.2} & \textbf{88.6} \\
\bottomrule
\end{tabular}
\end{table}

\begin{figure*}[t]
  \centering
  \includegraphics[width=\textwidth]{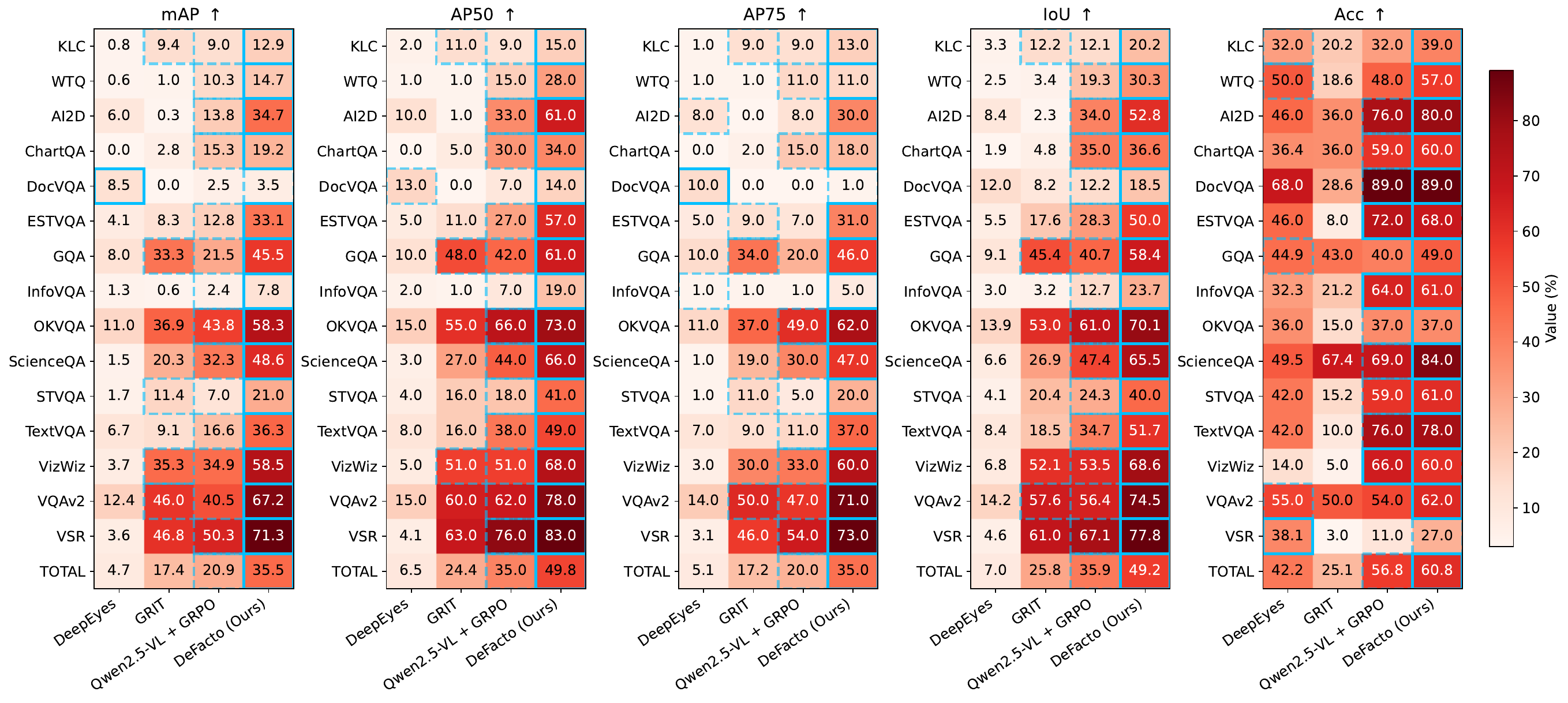}
  \caption{Heatmap visualization of faithful reasoning metrics across DeFacto-1.5K.}
  \label{fig:faithful_reasoning_heatmaps}
\end{figure*}

\begin{table*}[t]
\centering
\caption{Ablation study with component breakdown on representative benchmarks (accuracy, \%).}
\label{tab:ablation_components}
\small
\setlength{\tabcolsep}{2.2pt}
\renewcommand{\arraystretch}{1.0}
\begin{tabularx}{\textwidth}{@{}l c c c c c *{6}{>{\centering\arraybackslash}X}@{}}
\toprule
\textbf{Setting}
& \textbf{SFT}
& \textbf{CF Align.}
& \textbf{RandMask}
& \textbf{GRPO}
& \textbf{CF Reward}
& VQAv2 & OKVQA & SciQA & VSR & DVQA & TVQA \\
\midrule
(1) SFT (no CF)
& \checkmark & - & - & - & -
& 61.2 & 42.0 & 82.7 & 54.5 & 51.9 & 56.0 \\
(2) SFT + CF alignment
& \checkmark & \checkmark & \checkmark & - & -
& 66.5 & 55.7 & 84.7 & 53.7 & 84.3 & 73.0 \\
(3) GRPO (no CF reward)
& \checkmark & \checkmark & \checkmark & \checkmark & -
& 70.4 & 56.9 & \textbf{85.9} & 58.4 & 85.4 & 72.8 \\
\rowcolor{gray!20}
(4) DeFacto (full)
& \checkmark & \checkmark & \checkmark & \checkmark & \checkmark
& \textbf{72.1} & \textbf{61.7} & 83.6 & \textbf{71.0} & \textbf{94.0} & \textbf{82.9} \\
\midrule
\rowcolor{gray!10}
$\Delta$ (4) - (3)
& -- & -- & -- & -- & --
& +1.7 & +4.8 & -2.3 & +12.6 & +8.6 & +10.1 \\
\bottomrule
\end{tabularx}
\end{table*}

\subsection{Main results}  
\paragraph{Results on General VQA Benchmarks.}
Table~\ref{tab:gen_vqa} reports results on six widely used general-domain VQA benchmarks. Built upon the Qwen2.5-VL-7B backbone, \textsc{DeFacto} achieves the best performance on four out of six benchmarks, demonstrating consistent gains in both general reasoning and fine-grained visual understanding. Compared with the Qwen2.5-VL baseline, \textsc{DeFacto} improves accuracy from 67.0 to 72.1 on VQAv2 (+5.1), from 58.9 to 61.7 on OKVQA (+2.8), and from 53.2 to 63.9 on GQA (+10.7). Notably, \textsc{DeFacto} also delivers substantial improvements on perception-intensive datasets, reaching 83.6 on SciQA (+0.6), 61.4 on VizWiz (+7.3), and 71.0 on VSR (+7.4). Overall, \textsc{DeFacto} consistently surpasses representative thinking-with-images baselines (e.g., GRIT, DeepEyes, Visual-SR1, and ViCrop where applicable), establishing strong performance across all reported general VQA benchmarks.

\paragraph{Performance on Document and Text-centric Benchmarks.}
Table~\ref{tab:doc_scene_vqa} reports results on document VQA and scene text-centric benchmarks. \textsc{DeFacto} achieves strong overall performance, improving over Qwen2.5-VL on every benchmark, including DocVQA (+2.0), ChartQA (+7.7), InfoVQA (+7.6), DeepForm (+17.7), KLC (+2.4), WTQ (+1.4), STVQA (+3.3), TextVQA (+3.8), and AI2D (+6.6). \textsc{DeFacto} attains the best results on six out of nine benchmarks (DocVQA, ChartQA, InfoVQA, KLC, TextVQA, and AI2D), and remains competitive on the remaining three (DeepForm, WTQ, and STVQA). These results suggest that \textsc{DeFacto} effectively enhances evidence-grounded understanding for document and text-centric visual reasoning.

\paragraph{Faithful reasoning evaluation.}
Fig.~\ref{fig:faithful_reasoning_heatmaps} visualizes the evidence--answer consistency results on \textbf{DeFacto-1.5K} (100 samples per benchmark) using metric-wise heatmaps over datasets and methods. We report grounding quality and task performance measured by mAP/AP50/AP75, IoU, and answer accuracy. Following the standard detection interpretation, we adopt \emph{mAP} as the core faithfulness metric, which jointly reflects evidence localization and answer correctness: localization is quantified by IoU between the predicted evidence box and the ground-truth region, while the classification signal is defined by whether the predicted answer is correct.
For each dataset and metric, blue solid boxes indicate the best-performing method, while blue dashed boxes denote the second best.
Overall, \textsc{DeFacto} consistently yields the strongest evidence-grounded reasoning across benchmarks, achieving the best \textbf{TOTAL} scores (mAP 35.5, AP50 49.8, AP75 35.0, IoU 49.2) together with the highest accuracy (60.8). To dissect where these gains come from, we conduct a fine-grained failure-mode analysis on DeFacto-1.5K (Table~\ref{tab:faithfulness_analysis}). \textsc{DeFacto} substantially reduces \emph{Mislocalized \& Wrong} and \emph{Spurious Correct} compared with GRIT (43.5/11.6 vs.\ 21.5/5.6), and achieves the lowest \emph{Faithful but Wrong} rate (1.3). We further test counterfactual robustness by masking the human-annotated evidence or replacing the image entirely, and \textsc{DeFacto} achieves significantly higher abstention accuracy in both cases (64.1 and 61.4), confirming that its correctness genuinely depends on the visual evidence rather than language priors.

\begin{table}[h]
\centering
\caption{Region-level faithfulness analysis on DeFacto-1.5K.}
\label{tab:faithfulness_analysis}
\footnotesize
\setlength{\tabcolsep}{4pt}
\begin{tabular}{lccc}
\toprule
\textbf{Metric} & \textbf{DeepEyes} & \textbf{GRIT} & \textbf{Ours} \\
\midrule
Mislocalized \& Wrong $\downarrow$           & 23.5 & 43.5 & \textbf{21.5} \\
Spurious Correct $\downarrow$                & 11.0 & 11.6 & \textbf{5.6}  \\
Faithful but Wrong $\downarrow$              & 1.8  & 30.1 & \textbf{1.3}  \\
\midrule
Masked Evidence Abstention $\uparrow$        & 33.1 & 46.1 & \textbf{64.1} \\
Image Replacement Abstention $\uparrow$      & 40.3 & 46.3 & \textbf{61.4} \\
\bottomrule
\end{tabular}
\end{table}

\subsection{Ablation Study}

We compare four training settings on Qwen2.5-VL 7B: (i) \textbf{SFT (no CF)}: trained only on original data. (ii) \textbf{SFT (CF alignment)}: trained on original + counterfactual data, but counterfactuals are supervised only with the ``Unknown'' label, together with random-masking. (iii) \textbf{GRPO (no CF reward)}: “No CF reward’’ means that GRPO training uses only the first term of Eq.~\ref{eq:rans} (answer correctness) and the format reward. (iv) \textbf{DeFacto (full)}: our complete framework with all three rewards. In addition, we conduct a controlled ablation study on our \textbf{DeFacto-1.5K} benchmark to evaluate evidence--answer consistency under these training variants. Specifically, we report detection-style faithfulness metrics (mAP/AP50/AP75) computed against human-annotated evidence boxes, together with answer accuracy. 

\paragraph{Effect of Counterfactual Supervision.}
The first two rows in Table~\ref{tab:ablation_components} show that introducing counterfactual data with abstention alignment brings clear gains over standard SFT across most benchmarks. In particular, accuracy improves by +5.3\% on VQAv2 (61.2$\rightarrow$66.5) and +13.7\% on OKVQA (42.0$\rightarrow$55.7), with additional improvements on SciQA (+2.0\%) and substantial gains on text-centric tasks such as DVQA (DocVQA) (+32.4\%) and TVQA (TextVQA) (+17.0\%). These results suggest that counterfactual supervision effectively strengthens evidence--answer consistency and reduces spurious shortcut behaviors.

\paragraph{Effect of Reinforcement Learning.}
Compared with the GRPO baseline without counterfactual reward, \textsc{DeFacto} (CF reward + GRPO) achieves the strongest overall performance. As shown in Table~\ref{tab:ablation_components}, incorporating counterfactual rewards yields consistent improvements on five out of six benchmarks, including +1.7\% on VQAv2, +4.8\% on OKVQA, and a remarkable +12.6\% on VSR. Particularly large gains are observed on document and text-centric evaluations, with +8.6\% on DocVQA and +10.1\% on TextVQA. 

\begin{table}[ht]
\centering
\caption{Performance of Different Training Variants on the DeFacto-1.5K Validation Set.}
\label{tab:controlled_evaluation}
\scriptsize
\setlength{\tabcolsep}{4pt}
\renewcommand{\arraystretch}{1.05}
\begin{tabularx}{\columnwidth}{@{}>{\raggedright\arraybackslash}X c c c c@{}}
\toprule
Model Variant & mAP & AP50 & AP75 & Accuracy (\%) \\
\midrule
Qwen2.5-VL (Base)                
& 0.7  & 1.4  & 0.7  & 52.5 \\
SFT (no CF)         
& 0.0 & 0.0 & 0.0 & 53.8\\
SFT (CF alignment)  
& 2.9 & 4.4 & 1.7 & 57.0 \\
GRPO (no CF reward) 
& 20.9 & 35.0 & 20.0 & 56.8 \\
\rowcolor{gray!20}
\textbf{DeFacto (full)}          
& \textbf{35.5} & \textbf{49.8} & \textbf{35.0} & \textbf{60.8} \\
\bottomrule
\end{tabularx}
\end{table}

\paragraph{Faithfulness Evaluation.}
Table~\ref{tab:controlled_evaluation} compares the Qwen2.5-VL base model and several training variants on \textbf{DeFacto-1.5K}, where we evaluate evidence--answer consistency using detection-style faithfulness metrics (mAP/AP50/AP75) together with answer accuracy. In particular, our full framework reaches the highest mAP (35.5), AP50 (49.8), and AP75 (35.0), while also obtaining the highest accuracy (60.8). Compared to the base model and different component combinations, these results indicate that \textsc{DeFacto} most effectively improves evidence--answer consistency by jointly optimizing evidence selection and final prediction.

\section{Conclusion}
\label{sec:conclusion}
In this work, we introduced \textit{DeFacto}, a vision-language reasoning framework grounded in counterfactual supervision and designed to enforce region-faithful reasoning and abstention behavior when critical evidence is missing. To enable this paradigm, we proposed an automatic pipeline that leverages language model parsing, open-vocabulary detection, and OCR to mask question-relevant regions without manual annotation, and used it to construct a counterfactual dataset of about 100k images, together with a human-verified benchmark for evaluating evidence-grounded reasoning. Extensive experiments across diverse benchmarks demonstrate that \textit{DeFacto} consistently improves both answer accuracy and visual grounding faithfulness over strong baselines, while our ablation studies confirm the necessity of counterfactual training and region-level reward design. We note several limitations: our region proposal relies on publicly available detectors (e.g., RPN, OCR, and open-vocabulary models), which may introduce occasional errors but can be alleviated as stronger detectors become available; our counterfactual dataset, while sufficient for controlled experiments, remains modest compared to large-scale pretraining corpora; and our evaluation has so far focused on static images, leaving the extension to videos and temporal reasoning as an open direction. We believe these findings open new directions for integrating counterfactual supervision into multimodal reasoning systems, with potential extensions to video understanding and embodied AI.

\section*{Impact Statement}
This paper presents work whose goal is to advance the field of machine learning. There are many potential societal consequences of our work, none of which we feel must be specifically highlighted here.

\bibliography{example_paper}
\bibliographystyle{icml2026}



\end{document}